%% file: main.tex
\documentclass[10pt,logo,copyright]{giga-report}
\usepackage{parskip}
\usepackage{multirow}
\usepackage{marvosym}
\usepackage[numbers,sort&compress]{natbib}
\usepackage[nameinlink]{cleveref}
\crefname{equation}{Eq.}{Eqs.}
\crefname{figure}{Fig.}{Figs.}
\crefname{section}{Sec.}{Secs.}
\crefname{table}{Tab.}{Tabs.}
\newcommand{\reporttitle}{DriveDreamer-RSI: Recursive Self-Improvement of Driving Perception and Agentic Data Generation}
\title{\reporttitle}
\renewcommand{\today}{2026-11-23}
\newcommand{\reportauthors}{%
  Dingyi~Yao\textsuperscript{1,2}, Tingdong~Yu\textsuperscript{1},
  Guosheng~Zhao\textsuperscript{1}, Chaojun~Ni\textsuperscript{1},
  Xiaofeng~Wang\textsuperscript{1,2}, Jiming~Mao\textsuperscript{1},
  Guan~Huang\textsuperscript{1}, Lihui~Peng\textsuperscript{2\,\Letter},
  Zheng~Zhu\textsuperscript{1\,\Letter}}
\author{\parbox{0.76\linewidth}{\centering\reportauthors}}
\affil{\textsuperscript{1}GigaAI\qquad\textsuperscript{2}Tsinghua University\\
  \Letter\, Corresponding authors}
\hypersetup{pdftitle={\reporttitle},pdfauthor={\reportauthors}}

\begin{document}
\maketitle

\begin{abstract}
\input{sections/abstract}
\end{abstract}
\abscontent

\input{sections/introduction}
\input{sections/related_work}
\input{sections/method}
\input{sections/experiments}
\input{sections/conclusion}

\clearpage
\bibliographystyle{unsrtnat}
\bibliography{main}
\end{document}

%% file: sections/abstract.tex
Controllable generation can expand training data for driving perception, but visual realism alone does not ensure learning gains. Effective generation must adapt both what data to generate and how to construct them as the perception model evolves. In this paper, we present DriveDreamer-RSI, a framework for recursive self-improvement of driving perception and agentic data generation through feedback between perception learning and data construction. Its outer loop implements Gap-Guided Target Adaptation, generating labeled videos from current perception gaps and updating targets after training. Its inner loop uses a vision-language agent to iteratively edit scenes through observation and execution feedback. Across rounds, Learning-Feedback-Driven Strategy Refinement derives reusable construction advice from execution experience and downstream learning outcomes. On the nuScenes validation set, DriveDreamer-RSI surpasses the dataset-replay baseline by 4.09 mAP and 4.13 NDS points using approximately half the synthetic frames.

%% file: sections/introduction.tex
\section{Introduction}
\label{sec:intro}
\input{sections/teaser}

Training driving perception models relies on large-scale, high-quality annotated data. However, collecting and annotating real-world driving data requires substantial time and human effort, limiting the scalability of training data acquisition~\cite{liang2024aide,yao2026styleprofiling,yao2026astad}. Controllable generation offers a means of supplementing training data by adjusting scene content, object configurations, and environmental conditions~\cite{wang2024drivedreamer,gao2024magicdrive,wen2024panaceaplus,zhao2025drivedreamer2}. However, visual fidelity, controllability, and diversity alone do not guarantee the training value of synthetic data. This value must be assessed through actual improvements in downstream perception performance~\cite{huang2025subjectdrive,zeng2026dream4drive,song2025syntheticevaluation}. Effective data generation therefore requires aligning scene construction with the current perception model's learning needs.

As the perception model learns from additional data, its remaining capability gaps and their priorities change. Generation targets should therefore follow the model's current capability gaps, directing supervision toward unresolved weaknesses rather than the initial distribution~\cite{huang2024actgen,li2025gendataagent,ma2024delphi,ma2026correctad,lang2026embodirsi}. We refer to this mechanism as \emph{Gap-Guided Target Adaptation} (\cref{fig:teaser}(b)): updating generation targets according to the perception model's evolving capability gaps.

Once a generation target is identified, it remains necessary to decide how to construct the corresponding training data. Different scene edits can serve the same target, yet differ in their execution outcomes and benefits to perception learning~\cite{he2026langdrivectrl,zeng2026dream4drive}. Construction strategies should therefore draw on past attempts, combining execution feedback with downstream learning outcomes to continually improve subsequent construction decisions~\cite{cubuk2019autoaugment,ruiz2019learningtosimulate,kar2019metasim}. We refer to this mechanism as \emph{Learning-Feedback-Driven Strategy Refinement} (\cref{fig:teaser}(c)): deriving reusable strategy advice from execution experience and learning outcomes of past construction attempts, and continually updating memory to guide the agent's subsequent data construction.

Combining these two mechanisms requires improving both driving perception and the data generation process that supports its learning. Sustaining this joint improvement calls for a recursive self-improvement (RSI) mechanism, through which feedback from execution and learning in each round guides subsequent capability improvement and data construction~\cite{han2026physicalrsimechanisms,luo2026physicalrsi}. Recent work on embodied systems explores using execution feedback and learning experience to coordinate scene and task generation, training data acquisition, and policy updates~\cite{lang2026embodirsi,qin2026embodiedsmith,wang2026adarsi}.

To this end, we introduce \emph{DriveDreamer-RSI}, a framework for recursive self-improvement of driving perception that jointly adapts generation targets and construction strategies. Specifically, the framework couples two loops: an outer loop that generates labeled videos from current perception gaps and refreshes generation targets after perception training, and an inner loop that uses a vision-language agent to iteratively edit scenes based on observations and execution feedback. Across rounds, it revises and reuses construction experience from execution outcomes and downstream learning gains to guide subsequent scene edits. Our main contributions are:
\begin{itemize}
    \item We introduce DriveDreamer-RSI, a recursive self-improvement framework that integrates driving perception learning and agentic data generation.
    \item We develop Gap-Guided Target Adaptation to align generation targets with the perception model's evolving capability gaps.
    \item We develop Learning-Feedback-Driven Strategy Refinement to improve subsequent data construction strategies using past experience.
    \item On the nuScenes validation set, DriveDreamer-RSI achieves 45.98 mAP and 55.74 NDS for 3D object detection, surpassing the dataset-replay baseline by 4.09 and 4.13, respectively.
\end{itemize}

%% file: sections/teaser.tex
\begin{figure}[!b]
    \centering
    \includegraphics[width=0.70\linewidth]{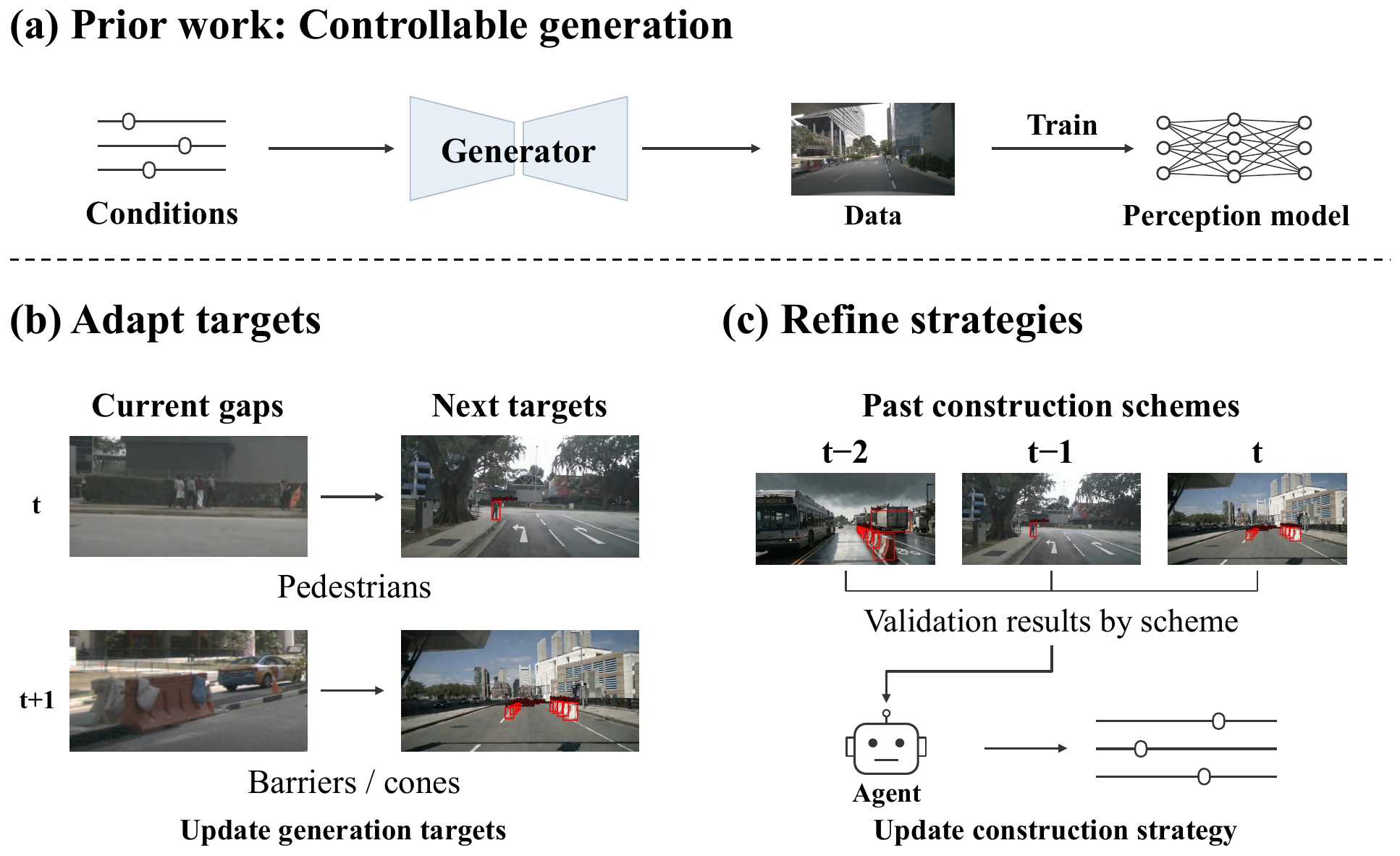}
    \captionsetup{font=small}
    \caption{\textbf{Target adaptation and strategy refinement in DriveDreamer-RSI.}
    \textbf{(a) Prior work.} Controllable driving generators use scene conditions to produce training data for perception models.
    \textbf{(b) Gap-Guided Target Adaptation.} The perception model's current capability gaps determine the next generation targets.
    \textbf{(c) Learning-Feedback-Driven Strategy Refinement.} The agent draws on past experience stored in memory to refine its data construction strategy.}
    \label{fig:teaser}
\end{figure}

%% file: sections/related_work.tex
\section{Related Work}
\label{sec:related_work}

\subsection{Driving Data Generation}
\label{sec:related_driving_generation}
DriveDreamer~\cite{wang2024drivedreamer}, MagicDrive~\cite{gao2024magicdrive}, and subsequent work~\cite{wen2024panacea,wen2024panaceaplus,lu2024wovogen,gao2025magicdrivev2} advance structural control and temporal and multiview consistency in driving scene generation. DriveDreamer-2~\cite{zhao2025drivedreamer2}, SimGen~\cite{zhou2024simgen}, and DriveArena~\cite{yang2025drivearena} further introduce language-driven trajectory control, simulator-conditioned generation, and closed-loop simulation, respectively. SubjectDrive~\cite{huang2025subjectdrive} and Dream4Drive~\cite{zeng2026dream4drive} explore synthetic data diversity and downstream perception gains, while ChatSim~\cite{wei2024chatsim} and LangDriveCTRL~\cite{he2026langdrivectrl} support language-driven editing. Building on these controllable generation and editing capabilities, we focus on adapting training data construction to the perception model's evolving learning needs.

\subsection{Adaptive Training Data}
\label{sec:related_adaptive_training_data}
General data-adaptation methods select or generate samples using model difficulty~\cite{shrivastava2016ohem,yoo2019learningloss,jain2023failuredirections,huang2024actgen,li2025gendataagent} and optimize augmentation or simulation configurations through distribution matching or downstream performance~\cite{cubuk2019autoaugment,ho2019pba,cheng2020ppba,ruiz2019learningtosimulate,kar2019metasim}. Driving-specific data engines connect model feedback to data acquisition and model updates across different tasks. AIDE~\cite{liang2024aide} targets open-world 2D object detection, linking novel-category discovery, image retrieval, automatic labeling, and continual detector updates. Delphi~\cite{ma2024delphi} and CorrectAD~\cite{ma2026correctad} use failure-driven data generation to improve end-to-end driving planning. Our work focuses on driving perception: we adapt generation targets using model feedback and further revise construction strategies across rounds using feedback from data construction and model training, allowing later construction to draw on experience from prior attempts.

\subsection{Agentic Generation and Recursive Improvement}
\label{sec:related_agentic_generation}
Research in robotics and general-purpose agents explores automated generation and feedback-driven improvement. Language agents generate simulation tasks, demonstrations, and skill-learning procedures~\cite{wang2024gensim,hua2025gensim2,wang2024robogen}. Eureka~\cite{ma2024eureka} revises rewards using policy-training feedback, while Reflexion~\cite{shinn2023reflexion} and Voyager~\cite{wang2024voyager} guide subsequent attempts through reflective memory and reusable skill libraries, respectively. EmbodiedSmith~\cite{qin2026embodiedsmith} uses simulation feedback to jointly refine scenes and task plans. EmbodiRSI~\cite{lang2026embodirsi} targets policy weaknesses with corrective experience. AdaRSI~\cite{wang2026adarsi} uses failure attribution and training experience to guide procedure revision and executor learning. ImageRSI~\cite{han2026imagersi} combines within-task repair with validated cross-task experience and judge updates. We extend these ideas to driving perception and data generation: we maintain and revise execution and learning experience separately to guide subsequent data construction.

%% file: sections/method.tex
\section{Method}
\label{sec:method}

\subsection{Overview and Recursive Formulation}
\label{sec:overview}

DriveDreamer-RSI couples a perception model $f_{\theta_t}$ with a generation agent guided by memory $\mathcal M_t$ through two nested loops (\cref{fig:overview}). In the outer loop, the current capability gaps $\mathcal F_t$ guide the construction of labeled synthetic videos $\widetilde{\mathcal D}_t$. Training on these videos and real data $\mathcal D_{\mathrm{real}}$ yields $f_{\theta_{t+1}}$, whose re-evaluation produces $\mathcal F_{t+1}$ for the next round.

\input{sections/framework}

Within data construction, the inner loop alternates agent actions with environment observations: the agent edits a scene, observes the resulting state and execution feedback, and revises its decisions until submission. The submitted scene is then used to generate videos.

Across rounds, construction records and downstream learning feedback revise the experience guiding future generation. The resulting recursion is
\begin{equation}
\begin{aligned}
\widetilde{\mathcal D}_t &= \mathcal G(\mathcal F_t;\mathcal M_t),\\
\theta_{t+1} &= \mathcal U(\theta_t,\mathcal D_{\mathrm{real}},\widetilde{\mathcal D}_t),\\
\mathcal M_{t+1} &= \mathcal R(\mathcal M_t,\mathcal T_t,\mathcal P_t),
\end{aligned}
\label{eq:recursive}
\end{equation}
where $\mathcal G$, $\mathcal U$, and $\mathcal R$ denote data construction, perception training, and experience revision, respectively; $\mathcal T_t$ contains construction records, and $\mathcal P_t$ contains evaluated construction schemes, each associated with its downstream evaluation results. The recursive state is $(\theta_t,\mathcal M_t)$: updated capability gaps redirect generation targets, while revised experience guides construction strategies.

\subsection{Adapting Generation Targets to Capability Gaps}
\label{sec:targets}

\subsubsection{Identifying Generation Targets}
\label{sec:identifying_generation_targets}
We evaluate $f_{\theta_t}$ on the validation set, rank diagnostic references by scene-level detection performance, and extract weak categories within each reference. A target $g_{t,i}$ comprises six surround-view images from an underperforming scene and weak-category hints. Scene performance identifies difficult contexts, while category-level analysis identifies the objects requiring additional supervision. The agent uses the reference to diagnose relevant conditions and the category hints to prioritize editing operations; these hints do not prescribe a fixed tool sequence. Repeating this analysis after perception training refreshes the target set as the learner changes.

\subsubsection{Matching Editable Scenes}
\label{sec:matching_editable_scenes}
A diagnostic reference provides the requirement and can be paired with multiple retrieved training scenes, each providing an editing substrate. Retrieval matches spatial layout and ego motion using
\begin{equation}
\begin{aligned}
\operatorname{sim}(x,y)={}&\lambda\operatorname{cos}(\ell(x),\ell(y))\\
&+(1-\lambda)\operatorname{cos}(r(x),r(y)),
\end{aligned}
\label{eq:retrieval}
\end{equation}
where $x$ is the diagnostic reference and $y$ is a candidate training scene. The layout descriptor $\ell$ summarizes spatial structure by spatially pooling the control latent while preserving its temporal dimension, then flattening and centering it. The trajectory descriptor $r$ summarizes ego motion through heading change, travel distance, curvature statistics, and turn category. The weight $\lambda$ balances structural and motion similarity.

Each matched training scene initializes a separate construction episode under the same generation target. For a given target--scene pair, we denote the initial state by $s_{t,i}^{0}$ and omit the additional match index for clarity. Structural and motion matching preserves relevant context without requiring identical appearance.

To prevent validation scenes from leaking into training data and support fair evaluation, the agent edits only scenes retrieved from the training set. Validation scenes serve solely as diagnostic references and are never directly edited to construct training samples.

\subsection{Interactive Scene Construction with Grounded Supervision}
\label{sec:construction}

\subsubsection{Agent--Environment Interaction}
\label{sec:agent_environment_interaction}
For each target--scene pair, the agent analyzes the diagnostic reference and edits the matched training scene. At step $k$, it selects an action based on the generation target $g$, the current observation $o^k$, and memory $\mathcal M_t$. The observation comprises the current scene image and textual execution feedback. Omitting the round and target indices, we write $g=g_{t,i}$ and $s^k=s_{t,i}^k$:
\begin{equation}
\begin{aligned}
a^k &\sim \pi_{\phi}(\cdot\mid g,o^k,\mathcal M_t),\\
(s^{k+1},o^{k+1}) &= \mathcal E(s^k,a^k),
\end{aligned}
\label{eq:interaction}
\end{equation}
where $\pi_\phi$ is the agent's construction policy and $\mathcal E$ executes $a^k$ to update the scene state and observation.

L1 memory maintains the working context of the current construction episode through its complete textual interaction history. It retains the initial diagnosis, subsequent tool calls and their parameters, and the environment's responses. These records capture the requested object categories, counts, and relative sides, as well as weather conditions. Execution feedback records whether an edit succeeds, fails, or is rejected, together with available reasons and quality-check outcomes. As interaction proceeds, new actions and observations are appended to the history and included in subsequent agent inputs. Together with the current scene image, this history allows the agent to relate the visible scene to previous edits, identify which attempts succeeded, and use earlier feedback when selecting the next operation or deciding to submit. The history is reset at the start of each new construction episode.

L2 and L3 memory provide reusable execution strategies and downstream learning evidence, respectively (see \cref{sec:revision}). Available actions are weather editing, object insertion, and submission. Episodes terminate on submission or budget exhaustion.

\subsubsection{Weather Editing}
\label{sec:weather_editing}
The six-view first-frame canvas serves as a shared state for weather editing and object insertion. For weather editing, the VLM agent selects a condition based on the diagnostic reference, and a text-guided image-editing model transforms the current canvas accordingly.

\subsubsection{Object Insertion}
\label{sec:object_insertion}
Object insertion follows a pick--draw--refit process (\cref{fig:pick-draw-refit}). The agent specifies the object category, count, and insertion side. Pick uses a VLM to propose a 2D anchor from the current front-view image augmented with road and occupancy cues, including previously inserted objects. A geometric solver combines this anchor with camera calibration, road geometry, and category-specific size priors to construct an initial 3D layout, whose projection defines an editing mask. Draw uses an image-editing model guided by the spatial layout and category-specific prompts to synthesize the objects in the current frame. Refit uses a second VLM to localize the rendered objects with tight 2D boxes in cropped views of the edited image. Geometric fitting then aligns the projected 3D boxes with these observations under scene and category constraints, producing annotations grounded in the rendered content. Geometric validity and per-box pixel-change checks determine edit acceptance.

\input{sections/pick_draw_refit}

\subsubsection{Training-Data Construction}
\label{sec:training_data_construction}
We pretrain a six-camera video generation model for high-resolution surround-view synthesis. Conditioned on the submitted first frame, refined scene-control sequences, and textual descriptions, the model generates videos. The resulting labeled synthetic videos form $\widetilde{\mathcal D}_t$, which is used for perception training.

This agent--environment formulation enables adaptive scene construction through iterative tool use and observation. Rather than committing to a fixed editing sequence, the agent observes intermediate images, scene states, and execution feedback to adjust tool selection, configuration, and action order, or decide when to submit. Since tools operate on a shared, evolving scene, subsequent editing operations can account for earlier edits and their interactions. Separating decision-making from execution also allows the agent to reason about which scene edits address a perception capability gap, while specialized tools handle spatial placement, image editing, and annotation refinement under scene constraints. This connects high-level construction intent to executable operations and supervision grounded in the rendered content. Beyond producing training data, the interaction trajectories expose the conditions, choices, and outcomes underlying construction, providing traceable evidence for the outer loop to revise strategies using execution outcomes and downstream learning feedback.

\subsection{Revising Construction Strategies through Learning Feedback}
\label{sec:revision}

\subsubsection{Construction Strategy}
\label{sec:construction_strategy}
The policy $\pi_\phi$ in \cref{eq:interaction} implements the construction strategy by selecting editing operations, tool configurations, action order, and submission decisions. Strategy refinement updates the experience in $\mathcal M_t$ that guides these decisions.

L1 memory provides working memory for the current episode, as detailed in \cref{sec:construction}. L2 memory captures whether and under which conditions an editing operation can be executed reliably; L3 memory captures whether an evaluated scheme benefits perception learning. L2 memory and L3 memory update when their respective feedback is available.

\subsubsection{Execution Experience (L2 Memory)}
\label{sec:execution_experience}
Episode records are grouped by condition, tool, and edit configuration. We aggregate execution outcomes, quality-check results, and failure and retry statistics, then summarize supported observations as reusable advice. Conditions and numerical evidence remain tied to the underlying records. These statistics describe associations within observed trajectories. L2 memory captures how an editing operation was executed and when its execution was reliable. For numerical feedback, we summarize each condition--operation group relative to the episodes used for reflection and retain its sample count.

\subsubsection{Learning Experience (L3 Memory)}
\label{sec:learning_experience}
After perception training, we collect downstream evaluation results for each construction scheme. Let $\mathcal P_t$ denote the set of schemes evaluated at round $t$, with each scheme linked to a record of its generation target, scene conditions, editing operations and their configurations, and corresponding perception results. These records include overall and category-level performance on the validation set, together with available reference results and observed improvements. These records provide the input for LLM-based experience synthesis.
The LLM jointly analyzes these records to identify patterns relating construction choices to learning outcomes. It examines which configurations improve targeted weak categories, the conditions under which gains occur, and differences between overall and category-level effects. It then distills these observations into conditional construction advice, specifying the relevant target and scene conditions, the recommended edit configuration, and the supporting evaluation results.
As perception learning changes the model's remaining weaknesses, new construction records and learning feedback are considered alongside existing L3 memory entries to refine their scope and advice.

\subsubsection{Revision and Reuse}
\label{sec:revision_and_reuse}
The LLM reviews new execution records and learning feedback together with the corresponding stored experience. It consolidates consistent observations, adjusts applicability conditions and recommendations when outcomes differ, and retires advice superseded by subsequent evidence. Revised entries retain their supporting records and can link to the entries they replace. L2 memory revision draws on execution outcomes, while L3 memory revision draws on construction descriptions and downstream perception results.

The resulting updates are
\begin{equation}
\begin{aligned}
\mathcal M_{t+1}^{(2)}&=\mathcal R_{\mathrm{exec}}(\mathcal M_t^{(2)},\mathcal T_t),\\
\mathcal M_{t+1}^{(3)}&=\mathcal R_{\mathrm{learn}}(\mathcal M_t^{(3)},\mathcal P_t),
\end{aligned}
\label{eq:experience}
\end{equation}
where $\mathcal R_{\mathrm{exec}}$ and $\mathcal R_{\mathrm{learn}}$ synthesize execution and learning experience, respectively. $\mathcal T_t$ contains the construction trajectories and their execution outcomes from round $t$, while $\mathcal P_t$ contains the evaluated construction schemes with their associated perception results. The updated L2 memory and L3 memory form a shared experience base that guides the strategy agent's decisions in the next round. This closes the feedback loop from execution and learning outcomes to construction strategy refinement.

%% file: sections/framework.tex
\begin{figure}[htbp]
    \centering
    \includegraphics[width=\linewidth]{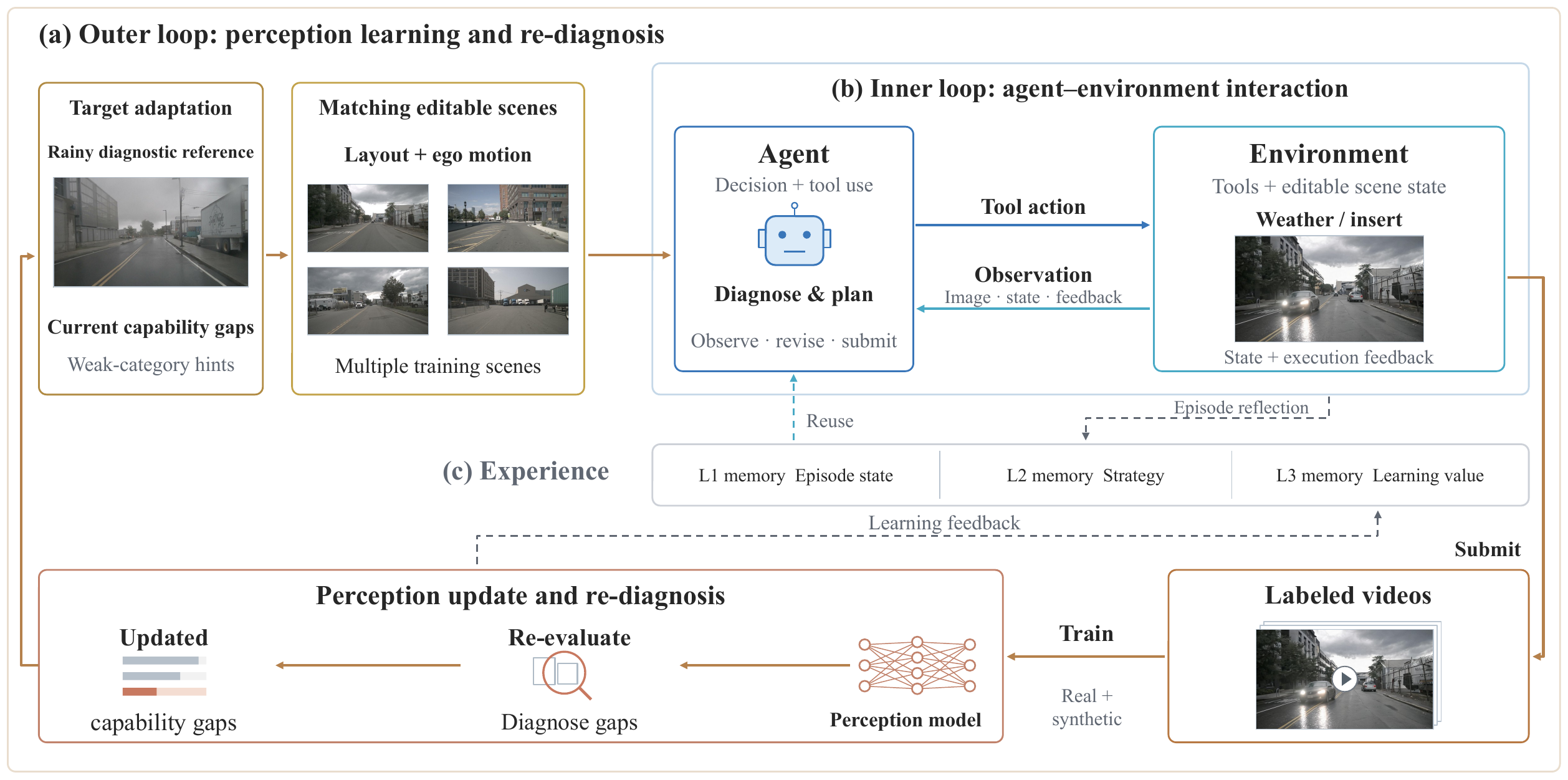}
    \caption{\textbf{Overview of DriveDreamer-RSI.}
    \textbf{(a) Outer loop.} Current capability gaps guide generation targets. Each diagnostic reference is matched to multiple training scenes by spatial layout and ego motion. Training on real and synthetic data updates the perception model. Re-evaluation reveals its remaining gaps for the next round.
    \textbf{(b) Inner loop.} For each target--scene pair, the agent diagnoses, plans, and edits with tools. Returned images, scene state, and execution feedback guide further edits until submission.
    \textbf{(c) Experience.} L1 memory tracks episode state, L2 stores strategies derived from execution, and L3 records downstream learning evidence.}
    \label{fig:overview}
\end{figure}

%% file: sections/pick_draw_refit.tex
\begin{figure}[htbp]
    \centering
    \includegraphics[width=0.87\linewidth]{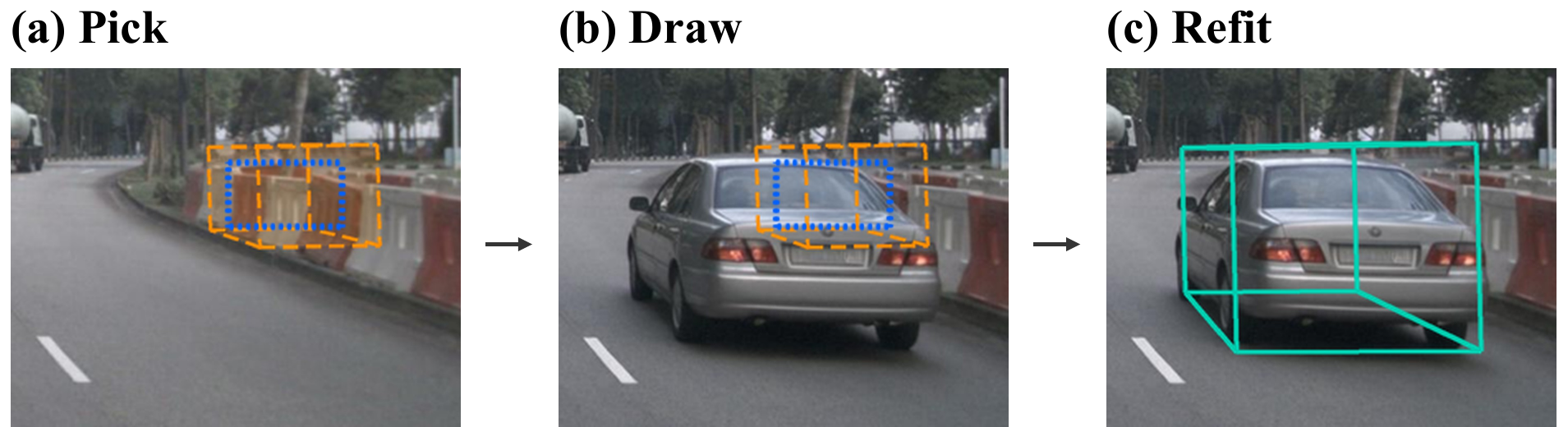}
    \caption{\textbf{Pick--Draw--Refit for object insertion.}
    \textbf{(a)} Pick selects a coarse 2D anchor (blue dotted rectangle) and builds a 3D prior (orange dashed box) and edit mask.
    \textbf{(b)} Draw renders the object, whose position and size may differ from the prior.
    \textbf{(c)} Refit localizes the object in 2D and fits its 3D box for supervision, shown in projection by solid teal lines.}
    \label{fig:pick-draw-refit}
\end{figure}

%% file: sections/experiments.tex
\section{Experiments}
\label{sec:experiments}

\subsection{Implementation Details}
\label{sec:implementation_details}

\subsubsection{Dataset and Evaluation}
\label{sec:dataset_evaluation}
We conduct six-camera 3D object detection experiments on nuScenes~\cite{caesar2020nuscenes}, using the official training and validation splits with 700 and 150 scenes, respectively. We evaluate detection performance using the official mAP and NDS metrics. The validation split is used for both generation target identification and performance evaluation, while all editable scenes used for synthetic data construction are retrieved from the training split.

\subsubsection{Perception Training}
\label{sec:perception_training}
We use StreamPETR~\cite{wang2023exploring} with a ResNet-50~\cite{he2016deep} backbone as the perception model and follow the training settings of DriveDreamer-2~\cite{zhao2025drivedreamer2}. Both real-only and real-plus-synthetic models train for 24 epochs. At each round, the model is trained on real data and synthetic data collected across rounds.

\subsubsection{Target Selection and Scene Retrieval}
\label{sec:target_selection_and_scene_retrieval}
Each round, we select 75 eligible validation scenes with the lowest mean AP over available categories. We provide up to four weak-category hints with the largest AP shortfalls relative to the same detector's full-validation category APs. For each reference, we retain 16 training clips after retrieval and filtering, with $\lambda=0.7$ in \cref{eq:retrieval}.

\subsubsection{Agent Configuration}
\label{sec:agent_configuration}
We use a GPT-6 model as the scene-construction agent. Given six-view scene images and weak-category hints, the agent edits the front-camera first frame through object-insertion and weather-editing tools. For object insertion, we use Seedream 5.0 Pro for image editing.

\subsubsection{Video Generation}
\label{sec:video_generation}
We train a six-camera surround-view video generator based on Helios~\cite{yuan2026helios}, conditioned on the edited first frame, scene-control sequences, and textual descriptions. Each camera view has a resolution of 1024×576.

The perception and video generation models are trained on 8 NVIDIA A800 GPUs. Further implementation details are provided in the supplementary material.

\subsection{Experimental Results}
\label{sec:experimental_results}


\subsubsection{Perception Performance and Data Efficiency}
\Cref{tab:perception_results} shows that at the same 448×256 detector input resolution, training with videos from our generator yields higher mAP and NDS than augmentation with DriveDreamer-2~\cite{zhao2025drivedreamer2}. Our generator also supports generation at 1024×576, where training with the full set of generated videos yields higher mAP and NDS than real-only training. Selecting scenes and object categories based on the detector's current capability gaps makes more effective use of synthetic data. In the first round, targeted generation outperforms the dataset-replay baseline on both metrics with a synthetic-to-real frame ratio of just $0.17:1$, versus nearly $1:1$ for the baseline. Later rounds target the updated detector's weaknesses. By round three, the lead over dataset-replay augmentation widens on both metrics, with about half as many synthetic frames in total.

\begin{table}[htbp]
\centering
\small
\setlength{\tabcolsep}{5pt}
\caption{\textbf{Downstream 3D detection on nuScenes validation.}
The resolution column reports detector input resolution, and Synthetic:Real denotes the ratio of synthetic to real training frames. Chunk-based generation discards incomplete scene tails, yielding ratios slightly below $1:1$ for full-set augmentation.}
\label{tab:perception_results}
\begin{tabular}{lccrr}
\toprule
Method / data & Resolution & Synthetic:Real & mAP$\uparrow$ & NDS$\uparrow$ \\
\midrule
Real only & 448×256 & 0:1 & 31.27 & 42.80 \\
DriveDreamer-2~\cite{zhao2025drivedreamer2} & 448×256 & 0.97:1 & 32.60 & 45.20 \\
Ours, dataset replay & 448×256 & 0.94:1 & 32.98 & 45.39 \\
\midrule
Real only & 1024×576 & 0:1 & 39.82 & 48.56 \\
Ours, dataset replay & 1024×576 & 0.94:1 & 41.89 & 51.61 \\
Ours, agent-in-the-loop (round 1) & 1024×576 & 0.17:1 & 44.17 & 53.98 \\
Ours, agent-in-the-loop & 1024×576 & 0.48:1 & \textbf{45.98} & \textbf{55.74} \\
\bottomrule
\end{tabular}
\end{table}

\subsubsection{Targeted Generation and Class-Specific Gains}
\label{sec:class_specific_gains}
Following the dataset replay protocol used in prior work~\cite{wang2024drivedreamer,zhao2025drivedreamer2,huang2025subjectdrive,wen2024panacea,wen2024panaceaplus}, which regenerates the full training set from first frames and scene conditions at a nearly $1:1$ synthetic-to-real frame ratio, our dataset-replay baseline gains 2.08 mAP points over real-only training.

Bicycles, motorcycles, and construction vehicles have the fewest annotated boxes in the training set. As shown in \cref{fig:class_specific_ap}, replay yields marginal AP gains for motorcycles and construction vehicles and lowers bicycle AP. In contrast, our targeted generation selects scenes and insertion categories based on the detector's current capability gaps, improving detection across all three classes.

\subsubsection{Detection Performance Across Recursive Rounds}
\label{sec:recursive_rounds}
\begin{figure}[htbp]
    \centering
    \begin{minipage}[t]{0.485\linewidth}
        \captionsetup{type=figure}
        \centering
        \includegraphics[height=0.55\linewidth]{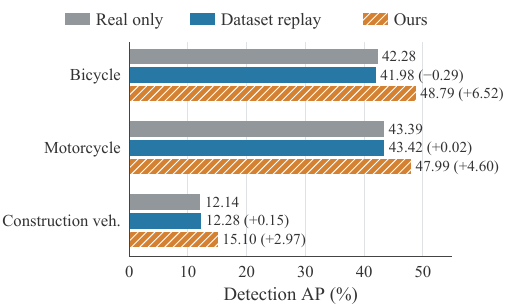}
        \caption{\textbf{Class-specific detection AP on nuScenes validation.} Parentheses indicate the change relative to real-only training.}
        \label{fig:class_specific_ap}
    \end{minipage}\hfill
    \begin{minipage}[t]{0.485\linewidth}
        \captionsetup{type=figure}
        \centering
        \includegraphics[height=0.55\linewidth]{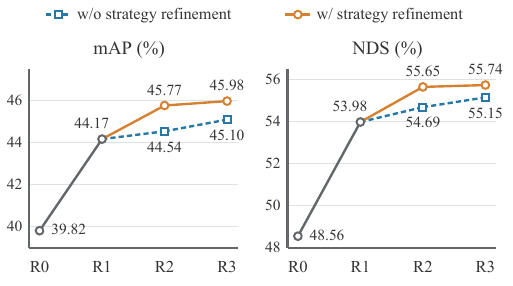}
        \caption{\textbf{Detection performance across recursive rounds with and without strategy refinement.} Both settings share the real-only detector (R0) and the first-round detector (R1).}
        \label{fig:recursive_rounds}
    \end{minipage}
\end{figure}

As shown in \cref{fig:recursive_rounds}, mAP and NDS improve in every round, both with and without strategy refinement. From round one to round three, the two settings gain 0.93--1.81 mAP points and 1.17--1.76 NDS points. After each training round, we re-evaluate the detector on the validation set, rank scenes by detection performance, and identify weak categories within the selected scenes. The continued gains suggest that adapting generation targets to the detector's evolving weaknesses provides useful training examples beyond the first round.

\subsection{Ablation Studies and Analysis}
\label{sec:ablations}

\subsubsection{Gap-Guided Target Adaptation}
\label{sec:target_adaptation_analysis}
The class-specific analysis in \cref{sec:class_specific_gains} shows that selecting generation targets based on detection weaknesses improves AP for categories that benefit little from dataset replay (\cref{fig:class_specific_ap}).

Beyond downstream detection performance, we examine how target guidance affects data construction by comparing random and gap-guided selection in the third round (\cref{tab:target_adaptation}). The Random baseline samples reference clips at random and provides no category hints. Gap-guided selects clips where the detector performs poorly and identifies the categories it struggles to detect in each scene. These hints give the agent a concrete editing target by specifying which objects to add to a difficult scene. With this guidance, the agent chooses object insertion more often, and the retained insertion clips contain more newly annotated boxes on average. This pattern suggests that pairing difficult scenes with their weak categories helps the agent translate capability gaps into edits that provide additional labeled instances in those scenes.

\begin{table}[htbp]
\centering
\small
\setlength{\tabcolsep}{6pt}
\caption{\textbf{Gap-guided target adaptation.} IE is the percentage of episodes with object insertion. BPC is the mean number of inserted boxes per retained insertion clip.}
\label{tab:target_adaptation}
\begin{tabular}{@{}lcc@{}}
\toprule
Method & IE (\%)$\uparrow$ & BPC$\uparrow$ \\
\midrule
Random & 34.41 & 6.76 \\
Gap-guided & \textbf{96.85} & \textbf{9.31} \\
\bottomrule
\end{tabular}
\end{table}

\subsubsection{Learning-Feedback-Driven Strategy Refinement}
\label{sec:strategy_refinement_analysis}

\begin{table}[htbp]
\centering
\small
\setlength{\tabcolsep}{6pt}
\caption{\textbf{Learning-Feedback-Driven Strategy Refinement.}
Comparison with and without strategy refinement. Actions: mean interaction steps per scene; IC: insertion calls; ISR: aggregate insertion success rate (\%).}
\label{tab:strategy_refinement}
\begin{tabular}{@{}clccc@{}}
\toprule
Round & Setting & Actions & IC & ISR$\uparrow$ \\
\midrule
\multirow{2}{*}{R2} & w/o strategy refinement & 4.07 & 1.66 & 68.5 \\
& w/ strategy refinement & 3.73 & 1.35 & \textbf{73.8} \\
\midrule
\multirow{2}{*}{R3} & w/o strategy refinement & 4.04 & 1.64 & 66.1 \\
& w/ strategy refinement & 3.51 & 1.14 & \textbf{86.9} \\
\bottomrule
\end{tabular}
\end{table}

\begin{figure}[htbp]
    \centering
    \includegraphics[width=\textwidth]{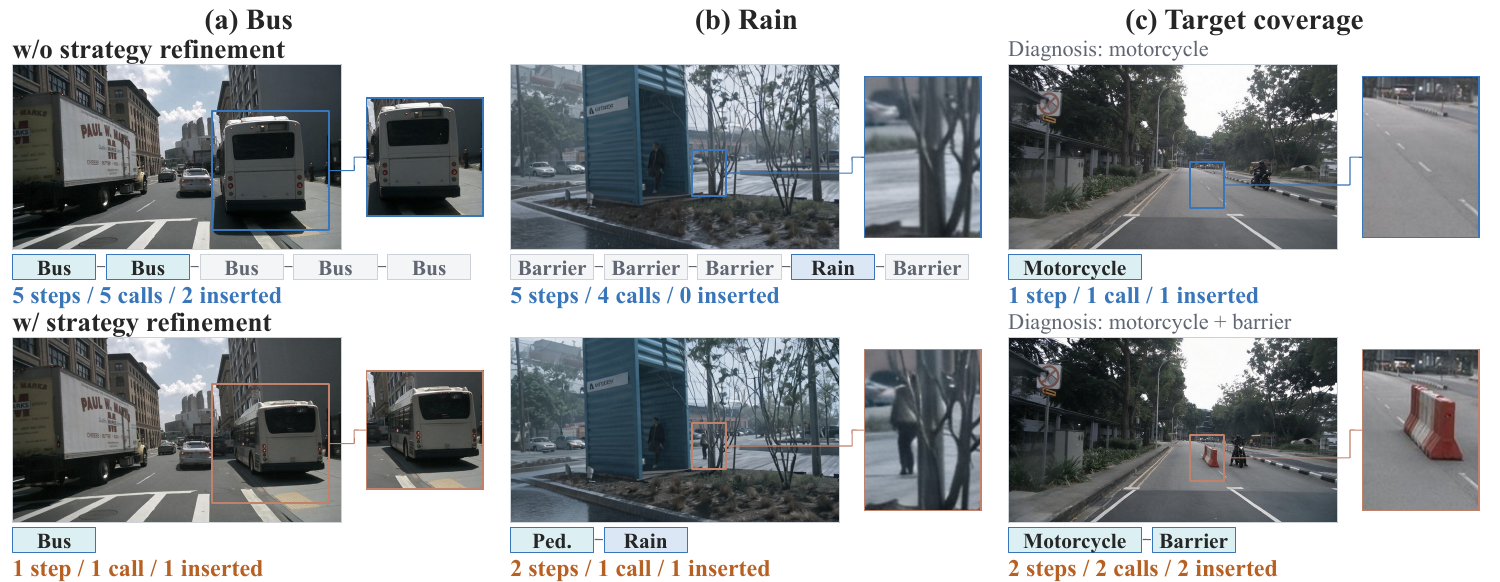}
    \caption{\textbf{Learning-Feedback-Driven Strategy Refinement in three matched third-round cases.}
    (a) The refined agent stops after one bus insertion. In the top row, the second bus occludes the first. (b) The refined agent inserts a pedestrian and applies rain. (c) Both agents insert a motorcycle. The refined agent also inserts a barrier identified in its diagnosis. Gray tiles indicate rejected insertion attempts, pale cyan tiles indicate successful insertions, and light blue tiles indicate weather edits. Steps count editing actions and calls count insertion attempts.}
    \label{fig:strategy_refinement_qualitative}
\end{figure}

\paragraph{Quantitative Analysis}
As shown in \cref{fig:recursive_rounds}, strategy refinement improves downstream detection performance in both R2 and R3 over the setting without refinement. This supports our view that adapting generation targets to capability gaps should be complemented by using past feedback to improve how those targets are realized. Execution experience informs editing feasibility, while learning feedback informs the training value of construction schemes. Together, they guide subsequent tool configurations, action order, and submission decisions, aligning data construction with both reliable execution and perception learning.

\Cref{tab:strategy_refinement} further examines strategy refinement at the execution level. Actions and Insertion Calls (IC) denote the mean numbers of interaction steps and insertion calls per scene, respectively; Insertion Success Rate (ISR) is the total number of successfully inserted instances divided by the total number of insertion calls, reported as a percentage. Strategy refinement reduces interaction steps and insertion calls while increasing ISR by 5.3 percentage points in R2 and 20.8 percentage points in R3. This indicates that accumulated experience helps the agent avoid ineffective attempts and perform insertions more reliably. Together with the detection gains in \cref{fig:recursive_rounds}, these results support the value of cross-round feedback in improving both the construction process and the training value of its outputs.

\paragraph{Qualitative Analysis}
\Cref{fig:strategy_refinement_qualitative} illustrates how Learning-Feedback-Driven Strategy Refinement adapts generation strategies to scene geometry and editing-tool capabilities using accumulated construction experience.

In the bus case \cref{fig:strategy_refinement_qualitative}(a), the agent without memory overlooks the drawing model's need for sufficient space and plausible placement, repeatedly attempting bus insertions that fail. Execution statistics and comparisons of construction schemes yield a strategy: continue insertion only when spatially feasible, and avoid adding instances without suitable placement. This strategy guides the agent to submit after one successful bus insertion, avoiding redundant insertions and subsequent failures.

In the rainy pedestrian case \cref{fig:strategy_refinement_qualitative}(b), barrier placement requires lane markings, yet the agent without memory repeatedly attempts insertion without usable markings. The refined strategy specifies: when barrier placement is infeasible, avoid repeated attempts and select another weak category for feasible insertion. Accordingly, the agent plans pedestrian insertion followed by a rain edit, successfully adding a target-category instance while preserving the weather objective.

In the target coverage case \cref{fig:strategy_refinement_qualitative}(c), the agent without memory focuses only on motorcycles, the category with the largest detection gap, and submits after inserting one. From the previous round's construction records and downstream detection results, L3 derives a strategy: attempt insertion for each diagnosed category still missing from the scene before submission. Given the same weak-category hints, the agent with memory targets both motorcycle and barrier, applying this coverage strategy to insert a barrier after the motorcycle and then submit.

\subsubsection{Interactive Scene Construction}
\label{sec:interactive_construction_analysis}

\begin{figure}[htbp]
    \centering
    \includegraphics[width=0.83\linewidth]{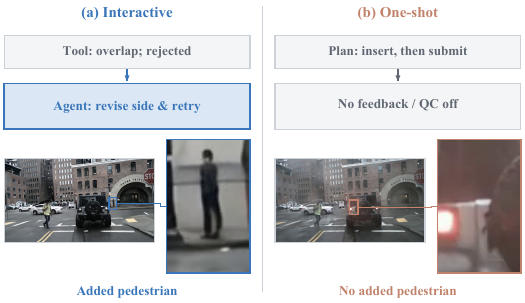}
    \caption{\textbf{Feedback-driven action revision.}
    \textbf{(a) Interactive editing.} The agent switches sides after a rejected insertion and successfully adds the pedestrian.
    \textbf{(b) One-shot editing.} The plan retains a box annotation for a pedestrian that was never inserted.}
    \label{fig:interactive_construction}
\end{figure}

\paragraph{Quantitative Analysis}
Interactive scene construction uses intermediate execution feedback to revise subsequent actions. Across 1,200 construction episodes in R1, 431 of 520 failed insertion actions were followed by another insertion attempt. Of these subsequent attempts, 345 retained the target category, including 329 that changed the requested side, while 86 switched categories. These results show that execution feedback informs the agent's subsequent decisions and leads to observable changes in its actions.

For comparison, we conduct an additional one-shot experiment that combines diagnosis and the complete editing plan into a single agent call. Tools then execute the plan without further feedback to the agent. Among 1,248 insertion attempts in this setting, 117 had a zero target-region change score but were still accepted. Completing a tool call therefore does not ensure that the target has been inserted, and retaining such edits with their insertion annotations can introduce erroneous supervision.

\paragraph{Qualitative Analysis}
\Cref{fig:interactive_construction} illustrates feedback-driven action revision. The first pedestrian insertion is rejected because the proposed placement overlaps an existing object. The agent responds by retaining the pedestrian target, revising the requested side, and retrying. The generated video then contains the added pedestrian. One-shot execution on the same scene produces no corresponding target but retains its insertion box, creating an image--annotation mismatch. This case shows that interactive scene construction enables the agent to revise failed edits using execution feedback, achieving target insertion while reducing erroneous supervision from image--annotation mismatches.

%% file: sections/conclusion.tex
\section{Conclusion}
\label{sec:conclusion}

We presented DriveDreamer-RSI, a recursive self-improvement framework connecting driving perception with agentic data generation. It adapts generation targets to evolving perception gaps, revises scene edits through execution feedback, and refines construction strategies using cross-round experience and learning outcomes. On nuScenes, it improves 3D detection over dataset-replay augmentation with approximately half as many synthetic frames. Strategy refinement further improves detection while reducing construction operations. Future work could explore SFT and RL for open-source VLM agents using accumulated experience and learning feedback, and extend the framework to additional tools and perception models.

%% file: main.bib
@inproceedings{wang2024drivedreamer,
  author = {Xiaofeng Wang and Zheng Zhu and Guan Huang and Xinze Chen and Jiagang Zhu and Jiwen Lu},
  title = {{DriveDreamer}: Towards Real-World-Drive World Models for Autonomous Driving},
  booktitle = {European Conference on Computer Vision (ECCV)},
  year = {2024},
  pages = {55--72}
}

@inproceedings{gao2024magicdrive,
  author = {Ruiyuan Gao and Kai Chen and Enze Xie and Lanqing Hong and Zhenguo Li and Dit-Yan Yeung and Qiang Xu},
  title = {{MagicDrive}: Street View Generation with Diverse {3D} Geometry Control},
  booktitle = {International Conference on Learning Representations (ICLR)},
  year = {2024}
}

@inproceedings{zhao2025drivedreamer2,
  author = {Guosheng Zhao and Xiaofeng Wang and Zheng Zhu and Xinze Chen and Guan Huang and Xiaoyi Bao and Xingang Wang},
  title = {{DriveDreamer-2}: {LLM}-Enhanced World Models for Diverse Driving Video Generation},
  booktitle = {AAAI Conference on Artificial Intelligence},
  year = {2025}
}

@inproceedings{zeng2026dream4drive,
  author = {Kai Zeng and Zhanqian Wu and Kaixin Xiong and Xiaobao Wei and Xiangyu Guo and Zhenxin Zhu and Kalok Ho and Lijun Zhou and Bohan Zeng and Ming Lu and Haiyang Sun and Bing Wang and Guang Chen and Hangjun Ye and Wentao Zhang},
  title = {Rethinking Driving World Model as Synthetic Data Generator for Perception Tasks},
  booktitle = {International Conference on Learning Representations (ICLR)},
  year = {2026}
}

@article{qin2026embodiedsmith,
  author = {Yikai Qin and Yifei Deng and Mingjian Liang and Wenxuan Song and Zepeng Lin and Zhiyi Jiang and Jiajun Fu and Qiao Sun and Huashuo Lei and Xicheng Gong and Jiayi Chen and Han Zhao and Shuanghao Bai and Pengxiang Ding and Pengwei Wang and Haoang Li},
  title = {{EmbodiedSmith}: Scaling Embodied Data through Recursive Self-Improvement Flywheel in Simulation},
  journal = {arXiv preprint arXiv:2610.07969},
  year = {2026}
}

@article{lang2026embodirsi,
  author = {Haoran Lang and Haotao Lu and Shiyu Sang and Haoyang Luo and Guo Chen and Qun Li and Jingyi Yu and Ye Shi and Jingya Wang},
  title = {{EmbodiRSI}: Recursive Self-Improvement for Data-Efficient Robot Adaptation},
  journal = {arXiv preprint arXiv:2609.38905},
  year = {2026}
}

@article{luo2026physicalrsi,
  author = {Meng Luo and Zechen Bai and Shanqing Xu and Yanlin Li and Kaixin Li and Mengchen Qian and Weifeng Lu and Ran Zhang and Yingtian Zou and Ziyang Luo and Wei Chen and Hongzhan Lin},
  title = {Toward Recursive Self-Improvement in Physical {AI}},
  journal = {arXiv preprint arXiv:2610.29677},
  year = {2026}
}

@inproceedings{wen2024panacea,
  author = {Yuqing Wen and Yucheng Zhao and Yingfei Liu and Fan Jia and Yanhui Wang and Chong Luo and Chi Zhang and Tiancai Wang and Xiaoyan Sun and Xiangyu Zhang},
  title = {{Panacea}: Panoramic and Controllable Video Generation for Autonomous Driving},
  booktitle = {IEEE/CVF Conference on Computer Vision and Pattern Recognition (CVPR)},
  year = {2024}
}

@inproceedings{lu2024wovogen,
  author = {Jiachen Lu and Ze Huang and Zeyu Yang and Jiahui Zhang and Li Zhang},
  title = {{WoVoGen}: World Volume-aware Diffusion for Controllable Multi-camera Driving Scene Generation},
  booktitle = {European Conference on Computer Vision (ECCV)},
  year = {2024}
}

@inproceedings{gao2025magicdrivev2,
  author = {Ruiyuan Gao and Kai Chen and Bo Xiao and Lanqing Hong and Zhenguo Li and Qiang Xu},
  title = {{MagicDrive-V2}: High-Resolution Long Video Generation for Autonomous Driving with Adaptive Control},
  booktitle = {IEEE/CVF International Conference on Computer Vision (ICCV)},
  year = {2025}
}

@inproceedings{zhou2024simgen,
  author = {Yunsong Zhou and Michael Simon and Zhenghao Peng and Sicheng Mo and Hongzi Zhu and Minyi Guo and Bolei Zhou},
  title = {{SimGen}: Simulator-conditioned Driving Scene Generation},
  booktitle = {Advances in Neural Information Processing Systems (NeurIPS)},
  year = {2024}
}

@inproceedings{huang2025subjectdrive,
  author = {Binyuan Huang and Yuqing Wen and Yucheng Zhao and Yaosi Hu and Yingfei Liu and Fan Jia and Weixin Mao and Tiancai Wang and Chi Zhang and Chang Wen Chen and Zhenzhong Chen and Xiangyu Zhang},
  title = {{SubjectDrive}: Scaling Generative Data in Autonomous Driving via Subject Control},
  booktitle = {AAAI Conference on Artificial Intelligence},
  year = {2025}
}

@inproceedings{wei2024chatsim,
  author = {Yuxi Wei and Zi Wang and Yifan Lu and Chenxin Xu and Changxing Liu and Hao Zhao and Siheng Chen and Yanfeng Wang},
  title = {Editable Scene Simulation for Autonomous Driving via Collaborative {LLM}-Agents},
  booktitle = {IEEE/CVF Conference on Computer Vision and Pattern Recognition (CVPR)},
  pages = {15077--15087},
  year = {2024}
}

@inproceedings{he2026langdrivectrl,
  author = {Yun He and Francesco Pittaluga and Ziyu Jiang and Matthias Zwicker and Manmohan Chandraker and Zaid Tasneem},
  title = {{LangDriveCTRL}: Natural Language Controllable Driving Scene Editing with Multi-modal Agents},
  booktitle = {European Conference on Computer Vision (ECCV)},
  year = {2026}
}

@inproceedings{shrivastava2016ohem,
  author = {Abhinav Shrivastava and Abhinav Gupta and Ross Girshick},
  title = {Training Region-based Object Detectors with Online Hard Example Mining},
  booktitle = {IEEE Conference on Computer Vision and Pattern Recognition (CVPR)},
  year = {2016}
}

@inproceedings{yoo2019learningloss,
  author = {Donggeun Yoo and In So Kweon},
  title = {Learning Loss for Active Learning},
  booktitle = {IEEE/CVF Conference on Computer Vision and Pattern Recognition (CVPR)},
  pages = {93--102},
  year = {2019}
}

@inproceedings{cubuk2019autoaugment,
  author = {Ekin D. Cubuk and Barret Zoph and Dandelion Man{\'e} and Vijay Vasudevan and Quoc V. Le},
  title = {{AutoAugment}: Learning Augmentation Strategies From Data},
  booktitle = {IEEE/CVF Conference on Computer Vision and Pattern Recognition (CVPR)},
  pages = {113--123},
  year = {2019}
}

@inproceedings{ho2019pba,
  author = {Daniel Ho and Eric Liang and Xi Chen and Ion Stoica and Pieter Abbeel},
  title = {Population Based Augmentation: Efficient Learning of Augmentation Policy Schedules},
  booktitle = {International Conference on Machine Learning (ICML)},
  pages = {2731--2741},
  year = {2019}
}

@inproceedings{cheng2020ppba,
  author = {Shuyang Cheng and Zhaoqi Leng and Ekin Dogus Cubuk and Barret Zoph and Chunyan Bai and Jiquan Ngiam and Yang Song and Benjamin Caine and Vijay Vasudevan and Congcong Li and Quoc V. Le and Jonathon Shlens and Dragomir Anguelov},
  title = {Improving {3D} Object Detection through Progressive Population Based Augmentation},
  booktitle = {European Conference on Computer Vision (ECCV)},
  year = {2020}
}

@inproceedings{ruiz2019learningtosimulate,
  author = {Nataniel Ruiz and Samuel Schulter and Manmohan Chandraker},
  title = {Learning to Simulate},
  booktitle = {International Conference on Learning Representations (ICLR)},
  year = {2019}
}

@inproceedings{kar2019metasim,
  author = {Amlan Kar and Aayush Prakash and Ming-Yu Liu and Eric Cameracci and Justin Yuan and Matt Rusiniak and David Acuna and Antonio Torralba and Sanja Fidler},
  title = {{Meta-Sim}: Learning to Generate Synthetic Datasets},
  booktitle = {IEEE/CVF International Conference on Computer Vision (ICCV)},
  year = {2019}
}

@inproceedings{jain2023failuredirections,
  author = {Saachi Jain and Hannah Lawrence and Ankur Moitra and Aleksander Madry},
  title = {Distilling Model Failures as Directions in Latent Space},
  booktitle = {International Conference on Learning Representations (ICLR)},
  year = {2023}
}

@inproceedings{huang2024actgen,
  author = {Tao Huang and Jiaqi Liu and Shan You and Chang Xu},
  title = {Active Generation for Image Classification},
  booktitle = {European Conference on Computer Vision (ECCV)},
  year = {2024}
}

@inproceedings{li2025gendataagent,
  author = {Zhiteng Li and Lele Chen and Jerone T. A. Andrews and Yunhao Ba and Yulun Zhang and Alice Xiang},
  title = {{GenDataAgent}: On-the-fly Dataset Augmentation with Synthetic Data},
  booktitle = {International Conference on Learning Representations (ICLR)},
  year = {2025}
}

@inproceedings{liang2024aide,
  author = {Mingfu Liang and Jong-Chyi Su and Samuel Schulter and Sparsh Garg and Shiyu Zhao and Ying Wu and Manmohan Chandraker},
  title = {{AIDE}: An Automatic Data Engine for Object Detection in Autonomous Driving},
  booktitle = {IEEE/CVF Conference on Computer Vision and Pattern Recognition (CVPR)},
  pages = {14695--14706},
  year = {2024}
}

@inproceedings{ma2026correctad,
  author = {Enhui Ma and Lijun Zhou and Tao Tang and Jiahuan Zhang and Junpeng Jiang and Zhan Zhang and Dong Han and Kun Zhan and Xueyang Zhang and Xianpeng Lang and Haiyang Sun and Xia Zhou and Di Lin and Kaicheng Yu},
  title = {{CorrectAD}: A Self-Correcting Agentic System to Improve End-to-end Planning in Autonomous Driving},
  booktitle = {AAAI Conference on Artificial Intelligence},
  year = {2026}
}

@inproceedings{wang2024gensim,
  author = {Lirui Wang and Yiyang Ling and Zhecheng Yuan and Mohit Shridhar and Chen Bao and Yuzhe Qin and Bailin Wang and Huazhe Xu and Xiaolong Wang},
  title = {{GenSim}: Generating Robotic Simulation Tasks via Large Language Models},
  booktitle = {International Conference on Learning Representations (ICLR)},
  year = {2024}
}

@inproceedings{hua2025gensim2,
  author = {Pu Hua and Minghuan Liu and Annabella Macaluso and Yunfeng Lin and Weinan Zhang and Huazhe Xu and Lirui Wang},
  title = {{GenSim2}: Scaling Robot Data Generation with Multi-modal and Reasoning {LLMs}},
  booktitle = {Conference on Robot Learning (CoRL)},
  pages = {5030--5066},
  year = {2025}
}

@inproceedings{wang2024robogen,
  author = {Yufei Wang and Zhou Xian and Feng Chen and Tsun-Hsuan Wang and Yian Wang and Katerina Fragkiadaki and Zackory Erickson and David Held and Chuang Gan},
  title = {{RoboGen}: Towards Unleashing Infinite Data for Automated Robot Learning via Generative Simulation},
  booktitle = {International Conference on Machine Learning (ICML)},
  year = {2024}
}

@inproceedings{ma2024eureka,
  author = {Yecheng Jason Ma and William Liang and Guanzhi Wang and De-An Huang and Osbert Bastani and Dinesh Jayaraman and Yuke Zhu and Linxi Fan and Anima Anandkumar},
  title = {{Eureka}: Human-Level Reward Design via Coding Large Language Models},
  booktitle = {International Conference on Learning Representations (ICLR)},
  year = {2024}
}

@inproceedings{shinn2023reflexion,
  author = {Noah Shinn and Federico Cassano and Ashwin Gopinath and Karthik Narasimhan and Shunyu Yao},
  title = {{Reflexion}: Language Agents with Verbal Reinforcement Learning},
  booktitle = {Advances in Neural Information Processing Systems (NeurIPS)},
  year = {2023}
}

@article{wang2024voyager,
  author = {Guanzhi Wang and Yuqi Xie and Yunfan Jiang and Ajay Mandlekar and Chaowei Xiao and Yuke Zhu and Linxi Fan and Anima Anandkumar},
  title = {{Voyager}: An Open-Ended Embodied Agent with Large Language Models},
  journal = {Transactions on Machine Learning Research},
  year = {2024}
}

@article{wang2026adarsi,
  author = {Peidong Wang and Ying Chang and Zhiming Ma and Xiaocui Yang and Shi Feng and Daqi Liu and Yuzhe Yang and Xu Cao and Chuang Wang and Jian Su and Daling Wang},
  title = {{AdaRSI}: On-Demand Recursive Self-Improvement for Embodied Agents},
  journal = {arXiv preprint arXiv:2610.20474},
  year = {2026}
}

@article{han2026imagersi,
  author = {Shuo Han and Zhilin Wang and Peng Ling and Wenjia Gong and Yang Liu and Guanjun Jiang and Yukun Cao and Xike Xie},
  title = {{ImageRSI}: Verification-Driven Recursive Self-Improvement for Agentic Image Generation},
  journal = {arXiv preprint arXiv:2610.32120},
  year = {2026}
}

@article{han2026physicalrsimechanisms,
  author = {Xiaofeng Han and Kaize Luo and Zenghuang Fu and Zichong Zhu and Zhe Feng and Changwei Wang and Rongtao Xu and Weiliang Meng and Xiaopeng Zhang},
  title = {Recursive Self-Improvement in Physical {AI}: Mechanisms, Evidence, and Evaluation},
  journal = {arXiv preprint arXiv:2610.38266},
  year = {2026}
}

@article{ma2024delphi,
  author = {Enhui Ma and Lijun Zhou and Tao Tang and Zhan Zhang and Dong Han and Junpeng Jiang and Kun Zhan and Peng Jia and Xianpeng Lang and Haiyang Sun and Di Lin and Kaicheng Yu},
  title = {Unleashing Generalization of End-to-End Autonomous Driving with Controllable Long Video Generation},
  journal = {arXiv preprint arXiv:2406.01349},
  year = {2024}
}

@article{wen2024panaceaplus,
  author = {Yuqing Wen and Yucheng Zhao and Yingfei Liu and Binyuan Huang and Fan Jia and Yanhui Wang and Chi Zhang and Tiancai Wang and Xiaoyan Sun and Xiangyu Zhang},
  title = {{Panacea+}: Panoramic and Controllable Video Generation for Autonomous Driving},
  journal = {IEEE Transactions on Circuits and Systems for Video Technology},
  volume = {36},
  number = {2},
  pages = {2247--2258},
  year = {2026},
  doi = {10.1109/TCSVT.2025.3601553}
}

@inproceedings{yang2025drivearena,
  author = {Xuemeng Yang and Licheng Wen and Tiantian Wei and Yukai Ma and Jianbiao Mei and Xin Li and Wenjie Lei and Daocheng Fu and Pinlong Cai and Min Dou and Liang He and Yong Liu and Botian Shi and Yu Qiao},
  title = {{DriveArena}: A Closed-loop Generative Simulation Platform for Autonomous Driving},
  booktitle = {IEEE/CVF International Conference on Computer Vision (ICCV)},
  year = {2025},
  pages = {26933--26943}
}

@inproceedings{yao2026styleprofiling,
  author = {Dingyi Yao and Xinyao Han and Ruibo Ming and Zhihang Song and Lihui Peng and Jianming Hu and Danya Yao and Yi Zhang},
  title = {A Style-Based Profiling Framework for Quantifying the Synthetic-to-Real Gap in Autonomous Driving Datasets},
  booktitle = {IEEE Intelligent Vehicles Symposium (IV)},
  year = {2026}
}

@inproceedings{yao2026astad,
  author = {Dingyi Yao and Xinqi Zhang and Lihui Peng and Jianming Hu and Danya Yao and Yi Zhang},
  title = {{ASTAD}: Asymmetric Style Transfer for Synthetic-to-Real Adaptation in Autonomous Driving},
  booktitle = {European Conference on Computer Vision (ECCV)},
  year = {2026}
}

@inproceedings{song2025syntheticevaluation,
  author = {Zhihang Song and Dingyi Yao and Ruibo Ming and Lihui Peng and Danya Yao and Yi Zhang},
  title = {Synthetic Dataset Evaluation Based on Generalized Cross Validation},
  booktitle = {IEEE International Conference on Imaging Systems and Techniques (IST)},
  year = {2025},
  doi = {10.1109/IST66504.2025.11268359}
}

@inproceedings{caesar2020nuscenes,
  author = {Holger Caesar and Varun Bankiti and Alex H. Lang and Sourabh Vora and Venice Erin Liong and Qiang Xu and Anush Krishnan and Yu Pan and Giancarlo Baldan and Oscar Beijbom},
  title = {{nuScenes}: A Multimodal Dataset for Autonomous Driving},
  booktitle = {IEEE/CVF Conference on Computer Vision and Pattern Recognition (CVPR)},
  year = {2020}
}

@inproceedings{he2016deep,
  author = {Kaiming He and Xiangyu Zhang and Shaoqing Ren and Jian Sun},
  title = {Deep Residual Learning for Image Recognition},
  booktitle = {IEEE Conference on Computer Vision and Pattern Recognition (CVPR)},
  year = {2016}
}

@inproceedings{wang2023exploring,
  author = {Shihao Wang and Yingfei Liu and Tiancai Wang and Ying Li and Xiangyu Zhang},
  title = {Exploring Object-Centric Temporal Modeling for Efficient Multi-View {3D} Object Detection},
  booktitle = {IEEE/CVF International Conference on Computer Vision (ICCV)},
  year = {2023}
}

@article{yuan2026helios,
  author = {Shenghai Yuan and Yuanyang Yin and Zongjian Li and Xinwei Huang and Xiao Yang and Li Yuan},
  title = {{Helios}: Real Real-Time Long Video Generation Model},
  journal = {IEEE Transactions on Pattern Analysis and Machine Intelligence},
  year = {2026},
  doi = {10.1109/TPAMI.2026.3750238}
}
